\documentclass{article} 
\usepackage[preprint]{neurips_2026}

\usepackage{amsmath,amsfonts,bm}

\def\eqref#1{equation~\ref{#1}}

\def\1{\bm{1}}

\DeclareMathAlphabet{\mathsfit}{\encodingdefault}{\sfdefault}{m}{sl}
\SetMathAlphabet{\mathsfit}{bold}{\encodingdefault}{\sfdefault}{bx}{n}

\usepackage{hyperref}
\usepackage{url}

\title{RLSplat: Resource-Linear Budget Control for Feed-forward 3D Gaussian Splatting}

\author{%
  Injae Kim\\
  KAIST\\
  \texttt{injae.kim@kaist.ac.kr} \\
  \And
  Minseok Joo\\
  Korea University\\
  \texttt{wlgkcjf87@korea.ac.kr} \\
  \And
  Jihwan Park\\
  KAIST\\
  \texttt{jseven7071@kaist.ac.kr} \\
  \And
  Sehyung Kim\\
  Korea University\\
  \texttt{skim129@korea.ac.kr} \\
  \And
  Minseong Bae\\
  KAIST\\
  \texttt{bms2002@kaist.ac.kr} \\
  \And
  Hyunwoo J. Kim\thanks{Corresponding author.} \\
  KAIST\\
  \texttt{hyunwoojkim@kaist.ac.kr} \\
}

\usepackage{amsmath}
\usepackage{graphicx}
\usepackage{wrapfig}
\usepackage{multirow}
\usepackage{textcomp, gensymb}
\usepackage{dsfont}
\usepackage{tabularx}
\usepackage[linesnumbered,ruled,vlined]{algorithm2e}
\usepackage{algpseudocode}
\usepackage{color}
\usepackage{colortbl}
\usepackage{amsfonts}
\usepackage{booktabs}       
\usepackage{multirow}
\usepackage{cleveref}
\usepackage{capt-of}

\crefname{figure}{Fig.}{Figs.}
\crefname{table}{Table}{Tables}
\crefname{section}{Sec.}{Secs.}
\crefname{equation}{Eq.}{Eqs.}

\usepackage{enumitem}
\usepackage{float}

\newcommand{\OURMODEL}{RLSplat }

\newcommand{\eg}{\textit{e.g.},\xspace}
\newcommand{\ie}{\textit{i.e.},\xspace}

\begin{document}

\maketitle

\begin{abstract}
Feed-forward 3D Gaussian Splatting (FF3DGS) reconstructs 3D scenes from images in a single forward pass, making it well-suited to applications that must operate within limited resource budgets. To further improve its practicality, recent budget-controllable FF3DGS methods allow the number of Gaussians to be adjusted at inference, so that the output meets a given storage limit or rendering-latency constraint. However, these methods overlook the cost of inference itself: to produce a representation with the target number of Gaussians, they first predict a dense set of Gaussians and then prune or select from it, requiring nearly constant latency and peak memory regardless of the budget. We therefore propose RLSplat, a resource-linear budget-controllable FF3DGS framework whose representation size, inference latency, and peak memory all decrease linearly as the Gaussian budget is reduced. Instead of pruning or selection, \OURMODEL learns a \emph{Gaussian allocation policy} that distributes the budget across local scene regions, so that exactly the requested number of Gaussians is generated. These Gaussians are then refined through \emph{resource-linear Gaussian decoding}, in which each Gaussian attends only to nearby Gaussians found via an efficient hierarchical neighbor search, so that the decoding cost decreases linearly with fewer Gaussians. In zero-shot evaluation, \OURMODEL outperforms prior budget-controllable methods at every budget, surpassing the strongest baseline even with 4$\times$ fewer Gaussians, while running up to 4.1$\times$ faster with 9.6$\times$ less peak memory at low budgets. Code is available at \url{https://github.com/mlvlab/RLSplat}.
\end{abstract}

\section{{Introduction}}
\label{sec:intro}

3D Gaussian Splatting (3DGS)~\citep{kerbl20233dgs} has become a leading representation for photorealistic, real-time novel view synthesis, supporting applications from AR/VR~\citep{qian2024gaussianavatars,jiang2024vrgs} to robotics~\citep{keetha2024splatam,matsuki2024gsslam} and digital content creation~\citep{tang2024dreamgaussian,chen2024gsgen}. Its explicit form further allows it to serve as a 3D scene representation for downstream models, such as 3D vision-language models~\citep{li2025scenesplat,thai2025splattalk}. Since per-scene optimization of 3DGS takes minutes to hours and requires many calibrated views, feed-forward 3DGS (FF3DGS) methods instead predict Gaussians in a single forward pass~\citep{charatan2024pixelsplat,chen2024mvsplat}, and recent variants even operate directly on uncalibrated images~\citep{jiang2025anysplat}. Owing to this fast and readily available reconstruction, FF3DGS has become a practical choice for applications that build 3D scene representations on the fly. 

\begin{figure}[t]
  \centering
  \includegraphics[width=0.99\linewidth]{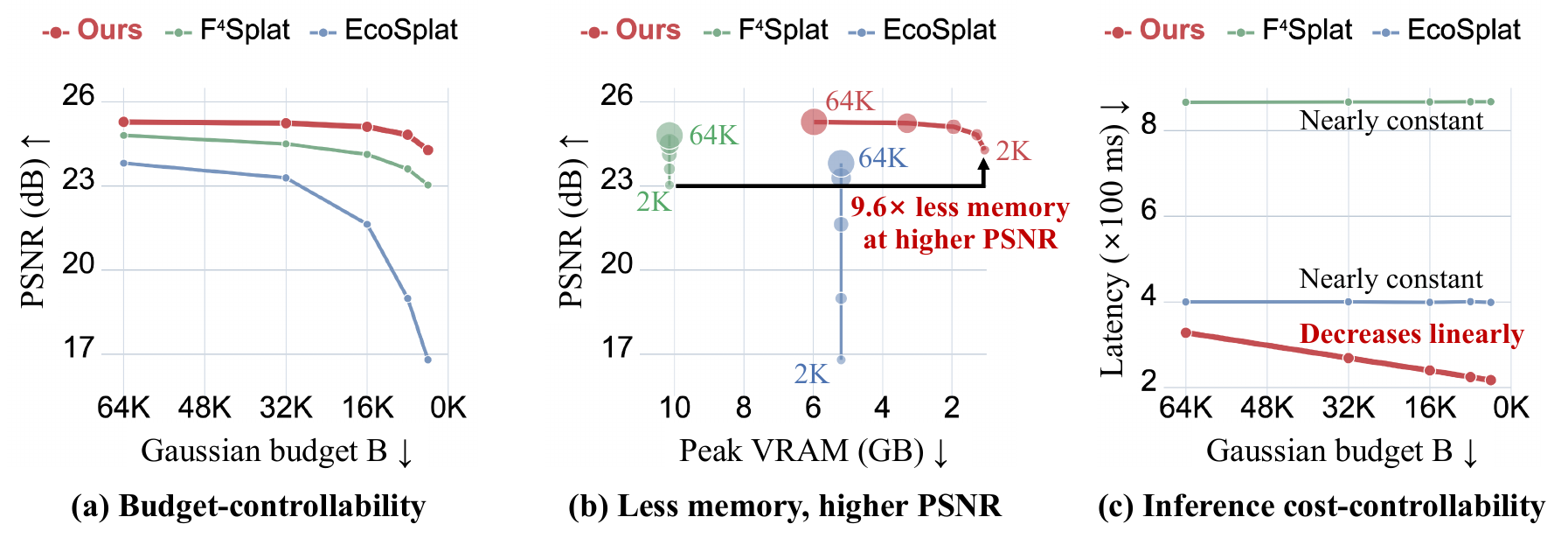}
  \caption{\textbf{Budget-controllable quality, memory, and latency trade-offs.} (a) RLSplat achieves the highest PSNR across the evaluated total Gaussian budgets \(B\). (b) At the 2K budget, RLSplat uses \(9.6\times\) less peak VRAM than F$^{4}$Splat while attaining higher PSNR. (c) As \(B\) is reduced, RLSplat’s inference latency decreases approximately linearly, whereas the latencies of F$^{4}$Splat and EcoSplat remain nearly constant. In axis tick labels, \(1\mathrm{K}=1{,}024\). }
  \label{fig:teaser}. 
  \vspace{-4mm}
\end{figure}

Early FF3DGS methods~\citep{charatan2024pixelsplat, chen2024mvsplat, ye2024nopo} simplify the problem by allocating a fixed number of Gaussians to each input pixel, which allows the feed-forward network to be trained effectively. However, this pixel-aligned design produces many redundant Gaussians, since Gaussians from overlapping views become increasingly redundant as more views are added. To address this redundancy, query-based methods~\citep{an2025c3g,itkin2026globalsplat} decode a fixed set of learnable query tokens into Gaussians. Yet neither approach can adjust the number of Gaussians without retraining. This lack of controllability makes it difficult to adapt to the diverse resource budgets encountered in real-world deployment.  To address this, recent works have proposed Gaussian budget-controllable methods. Given a Gaussian budget $B$, EcoSplat~\citep{park2025ecosplat} first predicts one Gaussian per pixel and keeps the $B$ Gaussians with the highest opacity. $\text{F}^4$Splat~\citep{kim2026f4splat} instead predicts Gaussians at multiple scales and selects the $B$ Gaussians. However, these methods overlook the computational cost of inference itself. Both methods first predict a much larger set of candidate Gaussians before reducing them to $B$, so their inference latency and peak memory remain nearly constant regardless of the budget, as shown in \cref{fig:teaser}(b) and (c).

We therefore propose RLSplat, a budget-controllable FF3DGS framework that is resource-linear: both its representation size and inference cost scale linearly with the budget $B$. Given uncalibrated images and a budget $B$, a geometry backbone encodes the images into anchor tokens, each representing a local 3D region of the scene. To obtain a compact representation, we formulate Gaussian generation as an allocation problem rather than pruning or selection: we predict how many Gaussians to assign to each anchor's region, summing to $B$. Since these discrete allocations are non-differentiable and can only be evaluated by rendering the resulting Gaussians, we model this prediction as a \emph{Gaussian allocation policy} and train it with policy optimization, using Group Sequence Policy Optimization (GSPO)~\citep{zheng2025gspo} with rendering quality as the reward.  Given the allocation, each anchor token spawns as many Gaussian tokens as its allocated count, yielding a total of $B$ Gaussian tokens. To efficiently decode these tokens into the final Gaussians, we introduce a \emph{resource-linear decoder}. It refines each Gaussian token through local cross-attention over its $K_{\rm{gs}}$ nearest Gaussian tokens, which we find with an efficient hierarchical neighbor search. Whereas global self-attention over all $B$ tokens costs $\mathcal{O}(B^2)$, each token here attends to only $K_{\rm{gs}}$ neighbors, so the decoding cost is $\mathcal{O}(BK_{\rm{gs}})$ and scales linearly with the Gaussian budget (\cref{fig:teaser}(c)). 
As a result, RLSplat achieves higher quality than prior budget-controllable methods at every budget (\cref{fig:teaser}(a)), while its peak memory and latency decrease with the budget. At 2K Gaussians, it achieves higher PSNR than F$^4$Splat while using 9.6$\times$ less peak memory (\cref{fig:teaser}(b)) and running 4.1$\times$ faster (\cref{fig:teaser}(c)).

In summary, our contributions are as follows:
\begin{itemize}
  \item We identify that existing budget-controllable FF3DGS methods control only the size of the output, not the cost of producing it, and introduce RLSplat, a resource-linear FF3DGS framework in which the number of Gaussians, inference latency, and peak memory all decrease linearly as the budget $B$ is reduced.
  \item We formulate Gaussian generation as an allocation problem and learn a \emph{Gaussian allocation policy} via policy optimization that places exactly $B$ Gaussians where the scene needs them, together with a \emph{resource-linear decoder} whose cost decreases linearly as the budget decreases through local cross-attention and an efficient hierarchical neighbor search.
  \item \OURMODEL consistently outperforms prior budget-controllable methods across all budgets. 
  At a budget of 2K Gaussians, it improves PSNR by 0.9 dB over F$^4$Splat while using 9.6$\times$ less peak memory and running 4.1$\times$ faster.
\end{itemize}

\section{Related Work}
\noindent\textbf{Feed-Forward 3D Gaussian Splatting.}
3D Gaussian Splatting (3DGS)~\citep{kerbl20233dgs} represents a scene with explicit Gaussian primitives and achieves real-time rendering, but requires time-consuming per-scene optimization from many calibrated views. Feed-forward 3DGS (FF3DGS) methods instead predict Gaussians in a single forward pass, from a single image~\citep{szymanowicz2024splatter} or from a few posed views~\citep{charatan2024pixelsplat,chen2024mvsplat,xu2025depthsplat,zhang2024gs}. Building on 3D foundation models~\citep{wang2024dust3r,leroy2024mast3r,wang2025vggt}, recent methods further remove the need for known camera poses~\citep{ye2024nopo,smart2024splatt3r,huang2025spfsplat,jiang2025anysplat}. Most of these methods predict one Gaussian per pixel or per voxel, so the number of Gaussians is determined by the input resolution and the number of views rather than by the scene content. This leads to redundant Gaussians in overlapping and textureless regions, and the number of Gaussians cannot be specified directly.

\noindent\textbf{Compact and Gaussian budget-controllable FF3DGS.}
For per-scene 3DGS, many works aim to represent a scene with fewer Gaussians without sacrificing quality, either by pruning redundant Gaussians~\citep{fan2024lightgaussian,lee2024compact,papantonakis2024reducing} or by deciding where to place Gaussians under a fixed or constrained budget during densification~\citep{fang2024mini,mallick2024taming,kheradmand20243d}. These methods, however, rely on iterative per-scene optimization. In FF3DGS, C3G~\citep{an2025c3g} and GlobalSplat~\citep{itkin2026globalsplat} decode a fixed set of learnable query tokens into compact sets of Gaussians, whose size is determined by the architecture. EcoSplat~\citep{park2025ecosplat} and $\text{F}^4$Splat~\citep{kim2026f4splat} instead accept a target count at inference, by keeping the Gaussians with the highest budget-conditioned opacity and by selecting one of multiple scales per region, respectively; both derive the output from a dense set of candidates. ZipSplat~\citep{veicht2026zipsplat}, developed concurrently with our work, controls the number of Gaussians by clustering visual tokens at inference. In contrast, we learn to allocate exactly the requested number of Gaussians and decode only those, so that our inference cost scales with the budget.

\noindent\textbf{Policy optimization.}
Policy gradient methods~\citep{williams1992reinforce,schulman2017ppo} optimize non-differentiable, discrete decisions from a scalar reward, and have been applied to generative vision models, \eg to align diffusion models with reward functions~\citep{black2024ddpo}. Group-relative variants such as GRPO~\citep{shao2024deepseekmath} and GSPO~\citep{zheng2025gspo} remove the need for a learned value function by normalizing rewards within a group of samples, and have become standard for fine-tuning large language models. In particular, GSPO defines the importance ratio from the likelihood of the whole sequence rather than of individual tokens, and clips it at the sequence level. We adopt this approach to learn Gaussian allocation, treating an allocation as a sequence of $B$ Gaussians and using rendering quality as the reward.
\section{Method}
\label{sec:method}
We present RLSplat, a feed-forward 3DGS framework whose Gaussian count and inference cost are both controlled by a single budget $B$. Rather than pruning or selecting from a dense set of Gaussians, \OURMODEL formulates Gaussian generation as allocation: a Gaussian allocation policy decides how many Gaussians each local region of the scene receives (\cref{sec:policy}), and a resource-linear decoder generates the allocated Gaussians and efficiently refines them with local cross-attention (\cref{sec:decoder}). We begin with an overview (\cref{sec:problem}) and conclude with the training procedure (\cref{sec:training}).

\begin{figure}[t]
  \centering
  \includegraphics[width=1.0\linewidth]{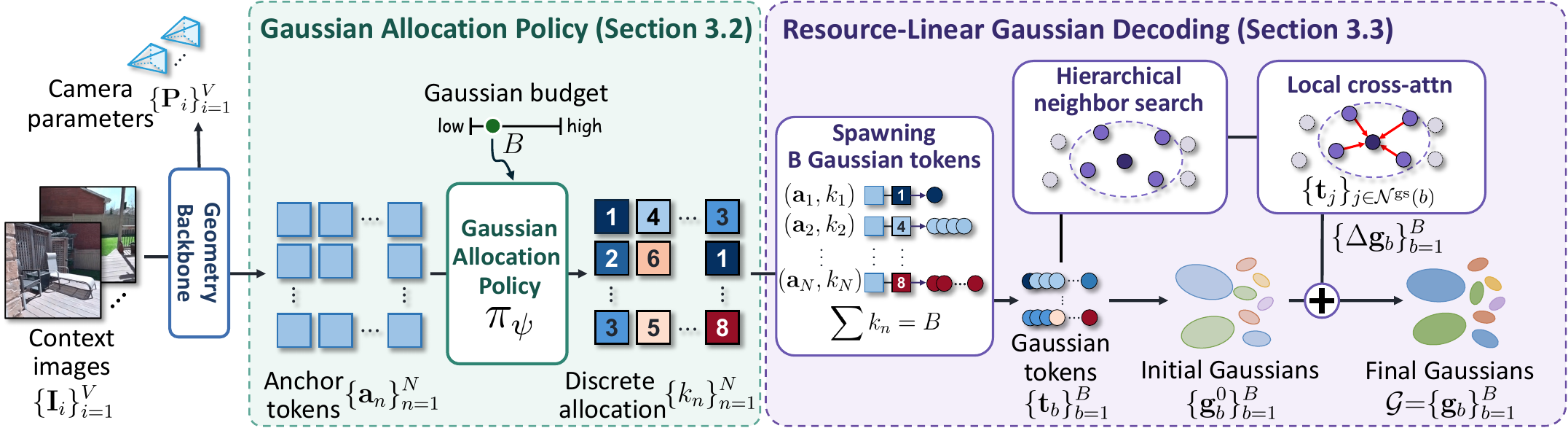}
  \vspace{-5mm}
  \caption{\textbf{Overview of RLSplat.}
  Given uncalibrated context images, the geometry backbone predicts camera parameters and extracts anchor tokens $\{\mathbf{a}_n\}$ representing local 3D regions. The Gaussian allocation policy distributes the budget $B$ across anchors, assigning $k_n$ Gaussians to each anchor with $\sum k_n = B$. Each anchor then spawns $k_n$ Gaussian tokens, which are decoded into $B$ Gaussians and locally refined using hierarchical neighbor search.}
  \label{fig:overview}
  \vspace{-5mm}
\end{figure}

\subsection{Overview of \OURMODEL}
\label{sec:problem}
\vspace{-2mm}

Given $V$ uncalibrated context images $\{\mathbf{I}_i\}_{i=1}^V$ and a target Gaussian budget $B$, \OURMODEL $\mathbf{F}_\theta$ predicts a set of 3D Gaussians $\mathcal{G}$ along with the context camera parameters $\{\mathbf{P}_i\}_{i=1}^V$:
\begin{equation}
  (\mathcal{G}, \{\mathbf{P}_i\}_{i=1}^V)
  = \mathbf{F}_\theta\big(\{\mathbf{I}_i\}_{i=1}^V, B\big),
  \qquad
  \mathcal{G} = \{\mathbf{g}_b\}_{b=1}^B.
  \label{eq:problem}
\end{equation}
Here, the 3DGS attribute $\mathbf{g}_b$ consists of a 3D center, opacity, scale, rotation, and color coefficients. 

As illustrated in \cref{fig:overview}, a geometry backbone~\citep{burzio2026dvlt} encodes the images into $N$ anchor tokens $\{\mathbf{a}_n\}_{n=1}^N$. These anchors serve as the foundational sites from which the 3D Gaussians are spawned. To use the limited budget effectively, \OURMODEL\ adaptively distributes the $B$ Gaussians across these anchors. Specifically, a Gaussian allocation policy assesses the local significance of each anchor to predict a discrete allocation $\mathbf{k} \in \mathbb{Z}_{\ge0}^N$ that strictly sums to the budget ($\sum k_n = B$), and we optimize this policy via reinforcement learning (\cref{sec:policy}). Next, a resource-linear decoder instantiates exactly $k_n$ Gaussian tokens at each anchor. It predicts an initial state for each Gaussian and efficiently processes them via a refinement module to produce the final set of $B$ Gaussians. This decoder's computational footprint scales linearly with the requested budget, as described in \cref{sec:decoder}.

\begin{wrapfigure}[25]{r}{0.4\textwidth}
  \vspace{-5pt}
  \centering
  \includegraphics[width=\linewidth]{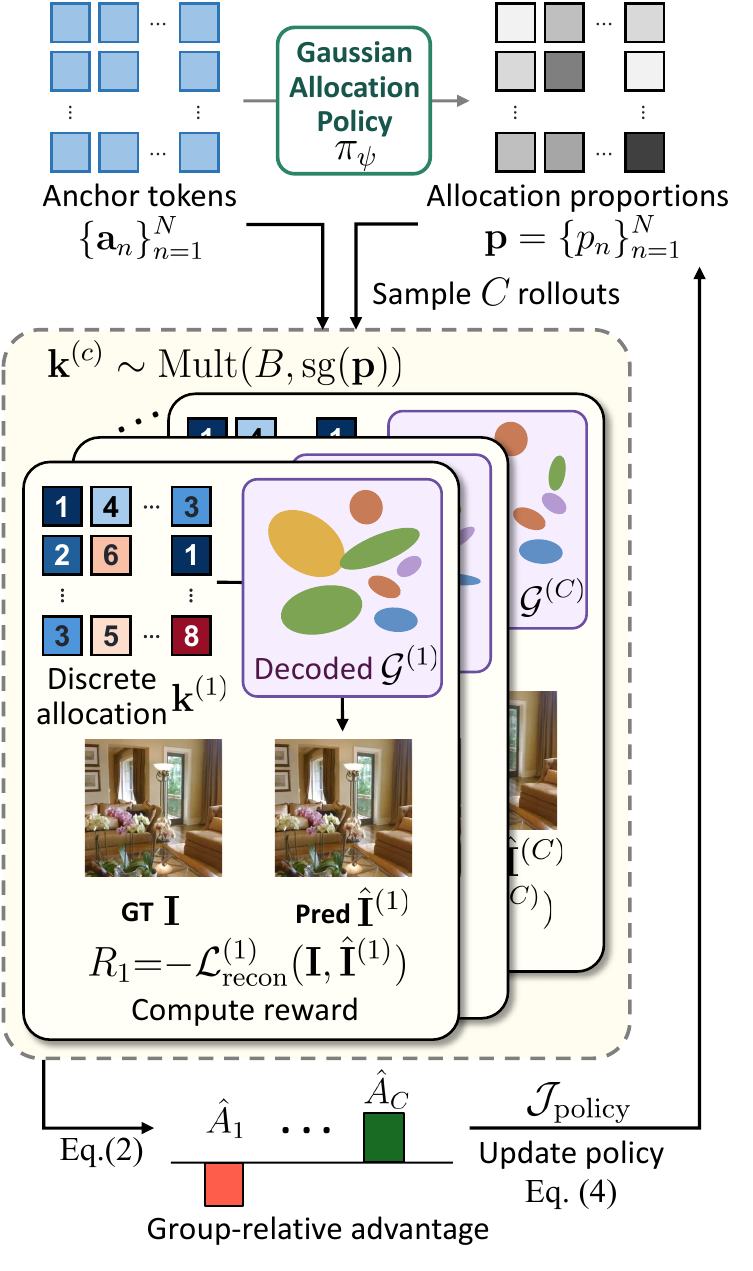}
  \vspace{-20pt}
  \caption{\textbf{Policy optimization.}}
  \label{fig:gspo}
\end{wrapfigure}
\vspace{-5mm}
\subsection{Gaussian Allocation Policy}
\vspace{-2mm}
\label{sec:policy}
To achieve a compact and high-fidelity representation, Gaussians must be densely allocated to the anchors that encode geometrically complex local 3D regions. We formulate this discrete allocation task as a reinforcement learning problem, parameterizing a policy $\pi_\psi(\mathbf{k} | \{\mathbf{a}_n\}_{n=1}^N, B)$ that maps the anchor tokens and target budget to a discrete allocation $\mathbf{k} = \{k_n\}_{n=1}^N$. Specifically, the policy model predicts allocation proportions $\mathbf{p} = \{p_n\}^N_{n=1}$, where $p_n$ represents the expected fraction of the Gaussian budget assigned to anchor $n$. The allocation process is modeled as drawing trials from a multinomial distribution, such that $\mathbf{k} \sim \operatorname{Mult}(B, \mathbf{p})$.

To optimize $\pi_\psi$ over these non-differentiable discrete allocations, we leverage Group Sequence Policy Optimization (GSPO)~\citep{zheng2025gspo}. At each training step, we draw a group of $C$ rollouts, \ie candidate allocations $\mathbf{k}^{(c)} \sim \pi_{\mathrm{old}} = \operatorname{Mult}(B, \operatorname{sg}(\mathbf{p}))$ for $c \in \{1, \dots, C\}$, from the current policy with gradients stopped by $\operatorname{sg}(\cdot)$. For each rollout $c$, we first generate the 3DGS representation $\mathcal{G}^{(c)}$. We then render an image $\hat{\mathbf{I}}^{(c)}$ from $\mathcal{G}^{(c)}$ at novel target views, computing the reconstruction loss $\mathcal{L}_{\mathrm{recon}}^{(c)}$ against the ground-truth $\mathbf{I}$ without gradient tracking. This loss directly serves as the reward signal for rendering quality, defined as $R_c = -\mathcal{L}_{\mathrm{recon}}^{(c)}$. Thanks to the highly efficient rasterization of 3DGS, evaluating multiple rollouts introduces minimal training overhead. We then compute the group-relative advantage $\hat{A}_c$ by standardizing the rewards within the group: 
\begin{equation}
    \hat{A}_c = \frac{R_c - \operatorname{mean}\left(\{R_{c'}\}_{c'=1}^C\right)}{{\operatorname{std}\left(\{R_{c'}\}_{c'=1}^C\right)}}.
\end{equation}

Following the GSPO framework, we view the allocation of Gaussians as a sequence of length $B$ and compute a length-normalized importance ratio $s_c$ for each rollout:
\begin{equation}
  s_c = \left(\frac{\pi_\psi(\mathbf{k}^{(c)})}{\pi_{\mathrm{old}}(\mathbf{k}^{(c)})}\right)^{\frac{1}{B}}
      = \exp\left(\frac{1}{B}\sum_{n=1}^N k^{(c)}_n \log \frac{p_n}{\operatorname{sg}(p_n)}\right).
  \label{eq:seq_ratio}
\end{equation}
The policy is then updated by maximizing the clipped objective
\begin{equation}
  \mathcal{J}_{\mathrm{policy}}
  = \mathbb{E}\left[\frac{1}{C}\sum_{c=1}^{C}
    \min\Big(s_c \hat{A}_c,\;
    \operatorname{clip}\big(s_c, 1-\epsilon, 1+\epsilon\big)\hat{A}_c\Big)\right],
  \label{eq:gspo}
\end{equation}
where $\epsilon$ is the clipping range, and the expectation is taken over scenes, budgets, and rollout groups. At inference time, we bypass stochastic sampling and deterministically convert the expected counts \(B\mathbf{p}\) into an integer allocation whose counts sum exactly to \(B\). Details of the rounding procedure are provided in \cref{supp:policy}.
\subsection{Resource-Linear Gaussian Decoding}
\label{sec:decoder}
Given a Gaussian allocation $\mathbf{k} = \{k_n\}_{n=1}^N$, the resource-linear decoder spawns exactly $B$ Gaussian tokens from the anchors and decodes them into the final Gaussians at a cost linear in $B$. We describe the token spawning, the local refinement with hierarchical neighbor search, and the regularizers that keep this search accurate.

\noindent\textbf{Gaussian token spawning.}
For each anchor $\mathbf{a}_n$, a token generator instantiates $k_n$ distinct Gaussian tokens, each conditioned on its slot index within the anchor (\cref{supp:decoder}):
\begin{equation}
  \{\mathbf{t}_b\}_{b=1}^{B} = \mathrm{TokenGen}\big(\{(\mathbf{a}_n, k_n)\}_{n=1}^N\big).
\end{equation}
We denote by $n_b$ the anchor that spawned token $b$. A head then predicts an initial Gaussian $\mathbf{g}_b^0 = \mathrm{Head}(\mathbf{t}_b)$ from each token, including its initial center $\bm{\mu}^0_b$.

\noindent\textbf{Local refinement with hierarchical neighbor search.}
Since the initial tokens lack spatial context, they must attend to neighboring Gaussians to refine their parameters. To avoid the prohibitive $\mathcal{O}(B^2)$ cost of global attention, we employ a local cross-attention mechanism over $L$ layers.
Anchors with $k_n=0$ are excluded from the hierarchical neighbor search. For each remaining anchor $n$, we compute the centroid $\bar{\bm{\mu}}_{n}$ of its initial Gaussians:
\begin{equation}
\bar{\bm{\mu}}_{n} = \frac{1}{k_n}\sum_{b:\, n_b=n} \bm{\mu}^0_b.
\end{equation}
To find neighbors efficiently, we search hierarchically: we find the $K_{\rm{anc}}$ nearest anchors, and then search for the $K_{\rm{gs}}$ nearest Gaussian tokens only among the Gaussians belonging to these local anchors:
\begin{equation}
    \begin{aligned}
      \mathcal{N}^{\mathrm{anc}}(n) &= \mathrm{KNN}_{K_{\rm{anc}}}\big(\bar{\bm{\mu}}_{n}; \{\bar{\bm{\mu}}_m:k_m>0\}\big), \\
      \mathcal{N}^{\mathrm{gs}}(b) &= \mathrm{KNN}_{K_{\rm{gs}}}\big(\bm{\mu}^0_b; \{\bm{\mu}^0_j : n_j \in \mathcal{N}^{\mathrm{anc}}(n_b)\}\big).
    \end{aligned}
\end{equation}
By restricting the search space, we keep the neighbor querying highly efficient. Both neighbor sets are computed once using the detached initial positions and shared across all $L$ layers. Each token $\mathbf{t}_b$ is then refined by $L$ layers of cross-attention over
its neighbors, each followed by a feed-forward layer:
\begin{equation}
  \mathbf{t}_b \leftarrow \mathbf{t}_b
    + \mathrm{CrossAttn}\big(\mathbf{t}_b,\,
      \{\mathbf{t}_j\}_{j \in \mathcal{N}^{\mathrm{gs}}(b)}\big).
  \label{eq:refine}
\end{equation}
Finally, a residual head predicts a correction $\Delta\mathbf{g}_b$ from $\mathbf{t}_b$, giving $\mathbf{g}_b = \mathbf{g}_b^0 + \Delta\mathbf{g}_b$.

\noindent\textbf{Locality regularizers.}
For this hierarchical search to be valid, a Gaussian's true spatial neighbors must actually reside within its parent's anchor neighborhood $\mathcal{N}^{\mathrm{anc}}(n_b)$. We enforce this using two regularizers. First, a Voronoi loss keeps each Gaussian closer to its own parent anchor's centroid $\bar{\bm{\mu}}_{n_b}$ than to any neighboring anchor's centroid $\bar{\bm{\mu}}_m$:
\begin{equation}
  \mathcal{L}_{\mathrm{vor}}
  = \frac{1}{B}\sum_{b=1}^{B} \sum_{m \in \mathcal{N}^{\mathrm{anc}}(n_b)}
    \max\left(0,\ \|\bm{\mu}^0_b - \bar{\bm{\mu}}_{n_b}\|
             - \|\bm{\mu}^0_b - \bar{\bm{\mu}}_m\|\right).
  \label{eq:vor}
\end{equation}
Second, an offset loss constrains the residual correction $\Delta\bm{\mu}_b$ to limit Gaussian drift:
\begin{equation}
  \mathcal{L}_{\mathrm{offset}}
  = \frac{1}{B}\sum_{b=1}^{B}
    \max\Big(0,\ \frac{\|\Delta\bm{\mu}_b\|}{d_{n_b}} - \delta\Big)^2,
  \label{eq:delta}
\end{equation}
where $d_n$ is the local anchor spacing, defined as the median distance from $\bar{\bm{\mu}}n$ to its $K{\rm anc}$ nearest anchors, and $\delta=0.5$ is a margin limiting Gaussian movement relative to the local anchor scale.

\subsection{{Training}}
\label{sec:training}

We train \OURMODEL in two stages: \emph{base training} trains the geometry backbone and decoder with random allocations, and \emph{policy training} introduces the Gaussian allocation policy and fine-tunes the entire model with policy optimization.

\noindent\textbf{Objectives.}
To supervise view synthesis, we apply a reconstruction loss $\mathcal{L}_{\mathrm{recon}}$, combining pixel-wise MSE and LPIPS between the rendered and ground-truth target views. For geometric alignment, a camera loss $\mathcal{L}_{\mathrm{cam}}$ supervises the translation, rotation, and field-of-view of the predicted context cameras. Additionally, a projection loss $\mathcal{L}_{\mathrm{proj}}$ reprojects each anchor's Gaussians toward their source image patch. Incorporating a cheirality penalty to prevent Gaussians from landing behind the camera, $\mathcal{L}_{\mathrm{proj}}$ effectively ties an anchor's allocation to a specific scene region and prevents arbitrary drift. These terms are combined with the locality regularizers ($\mathcal{L}_{\mathrm{vor}}$ and $\mathcal{L}_{\mathrm{offset}}$) previously introduced in \cref{sec:decoder}. Finally, during the policy training stage, we activate the policy loss $\mathcal{L}_{\mathrm{policy}} = -\mathcal{J}_{\mathrm{policy}}$ (\cref{eq:gspo}) to optimize the discrete allocations via GSPO. The overall training objectives are detailed in \cref{supp:training}. 

\noindent\textbf{Two-stage training.} We train \OURMODEL in two stages. In \emph{base training}, we train the encoder and decoder with random allocations, and sample budgets over the full supported range. This exposes the decoder to widely varying per-anchor counts, so that it learns to render diverse allocations before the policy is introduced. In \emph{policy training}, we add the policy head, initialized to output uniform probabilities (\cref{supp:policy}), and fine-tune the entire model with the policy loss $\mathcal{L}_{\mathrm{policy}}$. Since base training involves no policy, there is no reference policy to regularize toward, and unlike typical RL fine-tuning, we do not apply a KL penalty. During this stage, budgets are sampled with a skew toward the lower end of the range, where the choice of allocation matters most for rendering quality.
\section{Experiments}
\label{sec:exp}

\noindent\textbf{Datasets and evaluation.}
We train on RealEstate10K (RE10K)~\citep{zhou2018re10k} with $V \in \{2, 3, 4, 6, 8, 12, 24\}$ uncalibrated context views. To assess generalization, we evaluate the RE10K-trained model zero-shot on ACID~\citep{liu2021acid} (249 test scenes), which consists of aerial footage of coastal and natural scenes. We follow the evaluation protocol of $\text{F}^4$Splat~\citep{kim2026f4splat} at a resolution of $256\times256$ and report PSNR, SSIM, and LPIPS together with the number of Gaussians (\#GS). To measure inference cost, we report the latency and peak VRAM of Gaussian generation, excluding data loading, pose alignment, and rendering. All methods are run on a single RTX A6000 with a batch size of one. For each budget, we discard the first scene as warm-up, reset the peak-memory counter before each timed run, and report the mean over the test scenes.

\noindent\textbf{Baselines.} 
We mainly compare with \emph{budget-controllable} methods, EcoSplat~\citep{park2025ecosplat} and $\text{F}^4$Splat~\citep{kim2026f4splat}, which take a target Gaussian count as input at inference time. 
We evaluate all methods at the same requested Gaussian budgets, with particular focus on the low-budget regime. 
For a broader comparison, we also include \emph{budget-fixed} methods, which place Gaussians uniformly, either one per pixel (NoPoSplat~\citep{ye2024nopo}, SPFSplat~\citep{huang2025spfsplat}) or one per voxel (AnySplat~\citep{jiang2025anysplat}). Their Gaussian count is therefore determined by the input resolution and the number of views, or by the voxel size, rather than chosen by the user.

\noindent\textbf{Implementation details.}
In the base training stage, we train the geometry backbone and Gaussian decoder with random allocations for 40,000 steps. In the subsequent policy training stage, we train for an additional 5,000 steps. During training, the Gaussian budget is sampled from $[2048, 16384V]$, where $V$ is the number of input views. Unless otherwise noted, we use $K_{\rm{anc}}{=}K_{\rm{gs}}{=}16$ for Gaussian decoding in all experiments. Details are provided in \cref{supp:training,supp:ablation}.

\begin{table*}[t]
\centering
\setlength{\tabcolsep}{0.09cm}
\renewcommand{\arraystretch}{0.95}
\caption{\textbf{Budget-controllable methods on ACID (zero-shot from RE10K).}
Each method is evaluated at the same requested budget of 4K, 2K, and 1K Gaussians per context view; \#GS is the resulting total per scene (1K $=$ 1024). Best per column within each budget in bold.}
\label{tab:acid_budget_controllable}

\definecolor{oursbg}{RGB}{234,242,248}
\resizebox{\textwidth}{!}{
\begin{tabular}{lcccc cccc cccc}
\toprule
\multirow{2}{*}{\textbf{Method}}
& \multicolumn{4}{c}{\textbf{8 views}}
& \multicolumn{4}{c}{\textbf{16 views}}
& \multicolumn{4}{c}{\textbf{24 views}} \\
\cmidrule(lr){2-5} \cmidrule(lr){6-9} \cmidrule(lr){10-13}
& \#GS$\downarrow$ & LPIPS$\downarrow$ & SSIM$\uparrow$ & PSNR$\uparrow$
& \#GS$\downarrow$ & LPIPS$\downarrow$ & SSIM$\uparrow$ & PSNR$\uparrow$
& \#GS$\downarrow$ & LPIPS$\downarrow$ & SSIM$\uparrow$ & PSNR$\uparrow$ \\
\midrule

EcoSplat
& 32K & 0.300 & 0.669 & 23.29
& 64K & 0.282 & 0.689 & 23.86
& 96K & 0.284 & 0.688 & 23.81 \\
F$^4$Splat
& 32K & 0.240 & 0.718 & 24.50
& 64K & 0.218 & 0.743 & 25.09
& 96K & 0.211 & 0.753 & 25.31 \\
\rowcolor{oursbg}
\textbf{\OURMODEL}
& 32K & \textbf{0.211} & \textbf{0.756} & \textbf{25.24}
& 64K & \textbf{0.196} & \textbf{0.778} & \textbf{25.75}
& 96K & \textbf{0.192} & \textbf{0.787} & \textbf{25.94} \\

\midrule

EcoSplat
& 16K & 0.357 & 0.621 & 21.63
& 32K & 0.324 & 0.654 & 22.63
& 48K & 0.320 & 0.660 & 22.73 \\
F$^4$Splat
& 16K & 0.269 & 0.694 & 24.12
& 32K & 0.242 & 0.720 & 24.72
& 48K & 0.233 & 0.731 & 24.95 \\
\rowcolor{oursbg}
\textbf{\OURMODEL}
& 16K & \textbf{0.224} & \textbf{0.749} & \textbf{25.11}
& 32K & \textbf{0.205} & \textbf{0.773} & \textbf{25.63}
& 48K & \textbf{0.198} & \textbf{0.782} & \textbf{25.82} \\

\midrule

EcoSplat
& 8K & 0.456 & 0.533 & 18.99
& 16K & 0.412 & 0.575 & 20.03
& 24K & 0.402 & 0.587 & 20.28 \\
F$^4$Splat
& 8K & 0.307 & 0.664 & 23.61
& 16K & 0.276 & 0.690 & 24.21
& 24K & 0.263 & 0.702 & 24.45 \\
\rowcolor{oursbg}
\textbf{\OURMODEL}
& 8K & \textbf{0.254} & \textbf{0.734} & \textbf{24.82}
& 16K & \textbf{0.226} & \textbf{0.760} & \textbf{25.38}
& 24K & \textbf{0.215} & \textbf{0.771} & \textbf{25.58} \\

\bottomrule
\end{tabular}
}
\vspace{-10pt}
\end{table*}

\begin{table*}[t]
\centering
\setlength{\tabcolsep}{0.09cm}
\renewcommand{\arraystretch}{0.95}
\caption{\textbf{Budget-fixed methods on ACID (zero-shot from RE10K).}
These methods emit a fixed, dense set of Gaussians determined by the input. \OURMODEL outperforms them on all metrics while using substantially fewer Gaussians. Best per column in bold.}
\label{tab:acid_budget_fixed}

\definecolor{oursbg}{RGB}{234,242,248}
\resizebox{\textwidth}{!}{
\begin{tabular}{lcccc cccc cccc}
\toprule
\multirow{2}{*}{\textbf{Method}}
& \multicolumn{4}{c}{\textbf{8 views}}
& \multicolumn{4}{c}{\textbf{16 views}}
& \multicolumn{4}{c}{\textbf{24 views}} \\
\cmidrule(lr){2-5} \cmidrule(lr){6-9} \cmidrule(lr){10-13}
& \#GS$\downarrow$ & LPIPS$\downarrow$ & SSIM$\uparrow$ & PSNR$\uparrow$
& \#GS$\downarrow$ & LPIPS$\downarrow$ & SSIM$\uparrow$ & PSNR$\uparrow$
& \#GS$\downarrow$ & LPIPS$\downarrow$ & SSIM$\uparrow$ & PSNR$\uparrow$ \\
\midrule
NoPoSplat
& 512K & 0.268 & 0.672 & 22.84
& 1024K & 0.294 & 0.644 & 22.14
& 1536K & 0.307 & 0.630 & 21.78 \\
SPFSplat
& 512K & 0.215 & 0.736 & 24.89
& 1024K & 0.214 & 0.743 & 25.07
& 1536K & 0.218 & 0.744 & 25.00 \\
AnySplat
& 470K & 0.248 & 0.696 & 23.30
& 885K & 0.236 & 0.720 & 23.88
& 1259K & 0.234 & 0.727 & 24.04 \\
\rowcolor{oursbg}
\textbf{\OURMODEL}
& \textbf{32K} & \textbf{0.211} & \textbf{0.756} & \textbf{25.24}
& \textbf{32K} & \textbf{0.205} & \textbf{0.773} & \textbf{25.63}
& \textbf{24K} & \textbf{0.215} & \textbf{0.771} & \textbf{25.58} \\
\bottomrule
\end{tabular}
}
\vspace{-10pt}
\end{table*}

\begin{figure*}[t]
  \centering
  \includegraphics[width=1.0\linewidth]{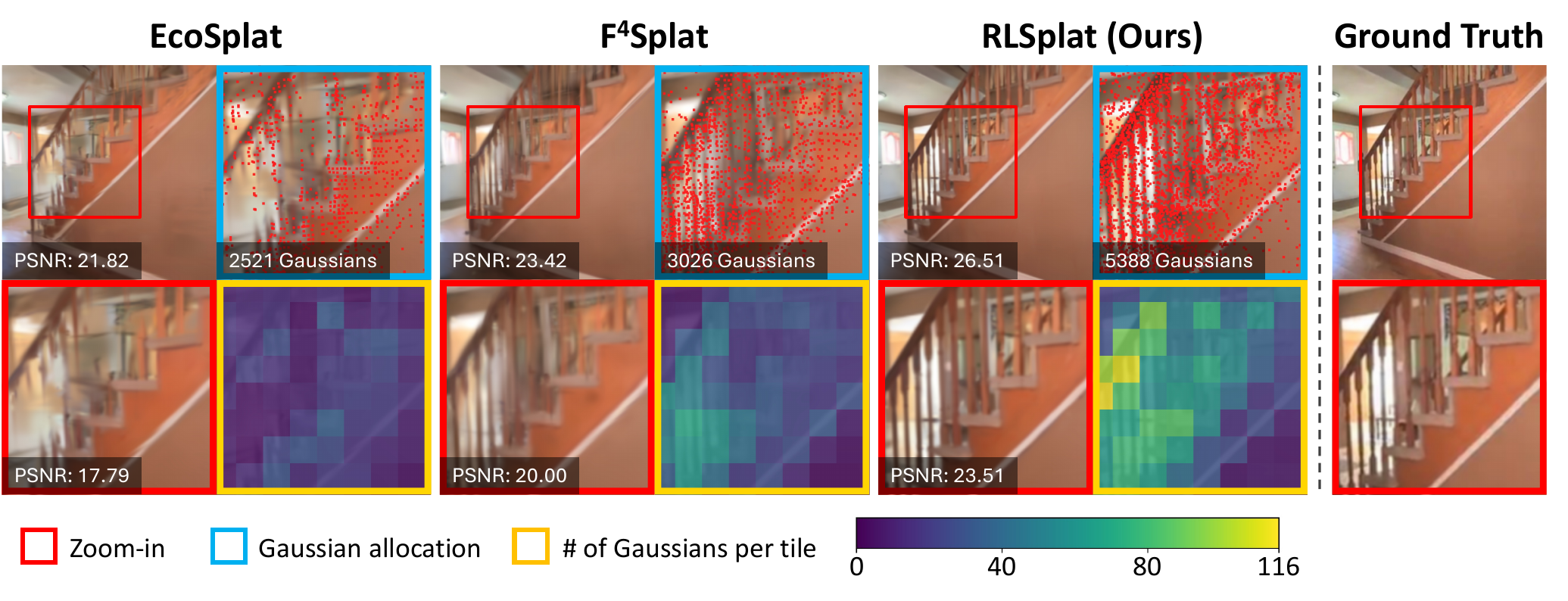}
    \vspace{-9mm}
    \caption{\textbf{Qualitative comparison under the same Gaussian budget.} All methods use 8K Gaussians on RE10K. For each method, we show the rendered view and PSNR, a zoomed region and its local PSNR, the Gaussian centers in the highlighted region (red), and a $16{\times}16$ tile-wise map of the number of Gaussians. RLSplat places more Gaussians around the thin stair balusters (5,388 vs.\ 3,026 for F$^4$Splat and 2,521 for EcoSplat), preserving the fine structure that the baselines blur. }
  \label{fig:qualitative}
  \vspace{-4mm}
\end{figure*}

\subsection{Novel-view Synthesis Results}
\label{sec:exp_results}

\Cref{tab:acid_budget_controllable} compares \OURMODEL with budget-controllable methods at matched budgets of 1K, 2K, and 4K Gaussians per context view. \OURMODEL achieves the best LPIPS, SSIM, and PSNR at every budget and view count. Importantly, its advantage becomes increasingly pronounced as the Gaussian budget decreases. For example, with 8 context views, the PSNR advantage over F$^4$Splat grows from 0.74 dB at 4K Gaussians per view to 0.99 dB at 2K and 1.22 dB at 1K. This trend is consistent across 16 and 24 context views, demonstrating that \OURMODEL preserves reconstruction quality particularly well under tight representation budgets. This robustness is consistent with how the final $B$ Gaussians are decoded: the baselines predict a dense set of Gaussians and then prune or select from it, whereas \OURMODEL directly decodes the requested number of Gaussians and refines them through resource-linear decoding.

For a broader comparison, \cref{tab:acid_budget_fixed} includes budget-fixed methods whose output size is determined by the input. Despite using substantially fewer Gaussians, \OURMODEL outperforms all of these methods on LPIPS, SSIM, and PSNR across all view counts. In particular, with 24 context views, \OURMODEL achieves 25.58 dB PSNR using only 24K Gaussians, compared with 25.00 dB for SPFSplat using 1536K Gaussians, corresponding to 64$\times$ fewer Gaussians.

\cref{fig:qualitative} shows how each method uses the same total budget of 8K Gaussians on RE10K. In the highlighted stair region, EcoSplat and F$^4$Splat retain 2,521 and 3,026 Gaussians, respectively, whereas \OURMODEL allocates 5,388 Gaussians to the thin, high-frequency structure. The tile-wise allocation maps further show that \OURMODEL concentrates its budget around the stair balusters while spending fewer Gaussians on the comparatively smooth wall. This targeted allocation leads to substantially better reconstruction of the thin structures: the local PSNR of the highlighted region increases from 17.79 dB for EcoSplat and 20.00 dB for F$^4$Splat to 23.51 dB for \OURMODEL. The improvement is also reflected in the full-view PSNR (21.82, 23.42, and 26.51 dB, respectively). These results illustrate that the learned allocation policy uses a fixed Gaussian budget more effectively by concentrating capacity where fine geometric and appearance details require it.
\vspace{-3mm}
\subsection{Ablation Studies and Analysis}
\vspace{-3mm}
\label{sec:exp_ablation}
Unless otherwise noted, all experiments in this section are evaluated on ACID with 8 views at a total budget of $B{=}8{,}192$ (1,024 Gaussians per view). Ablations that require retraining use the shortened schedules in \cref{supp:ablation}.

\begin{wraptable}[11]{r}{0.45\textwidth}
  \centering
  \footnotesize
  \setlength{\tabcolsep}{0.15cm}
  \vspace{-6mm}
  \caption{\textbf{Architecture and loss ablations.}
  $^\dagger$Without the projection loss, the model fails to reconstruct the scene.}
  \label{tab:ablation_arch}
  \begin{tabular}{lccc}
    \toprule
    Variant & LPIPS $\downarrow$ & SSIM $\uparrow$ & PSNR $\uparrow$ \\
    \midrule
    w/o $\mathcal{L}_{\mathrm{offset}}$               & 0.394 & 0.630 & 22.31 \\
    w/o $\mathcal{L}_{\mathrm{offset}}$, $\mathcal{L}_{\mathrm{vor}}$       & 0.479 & 0.547 & 20.33 \\
    \midrule
    w/o $\mathcal{L}_{\mathrm{proj}}$$^\dagger$ & 0.731 & 0.007 & 5.80 \\
    w/o local refinement          & 0.425 & 0.607 & 21.88 \\
    \midrule
    \textbf{Full model}                    & \textbf{0.353} & \textbf{0.643} & \textbf{22.84} \\
    \bottomrule
  \end{tabular}
\end{wraptable}
\noindent\textbf{Architecture and objectives.}
\Cref{tab:ablation_arch} ablates the locality regularizers, the projection loss, and the local refinement module. These variants change the base training; each is trained from scratch with the shortened pretraining schedule of \cref{supp:ablation}.
Removing the offset loss lowers PSNR by 0.53\,dB, as refinement then moves Gaussians far from the initial positions from which their neighborhoods are computed. Removing the Voronoi loss as well causes a much larger drop of 2.51\,dB, since Gaussians also leave their anchor's cell and the hierarchical search no longer retrieves their true neighbors. Without the projection loss, training fails entirely, since nothing ties an anchor's Gaussians to the image region it was encoded from. Removing the local refinement module lowers PSNR by 0.96\,dB, since each Gaussian is then predicted from its own token alone and cannot be made consistent with its neighbors.

\begin{figure}[t]
    \begin{minipage}[t]{0.48\columnwidth}
        \centering
        \includegraphics[trim={2pt 6pt -10pt 0}, clip, width=\linewidth]{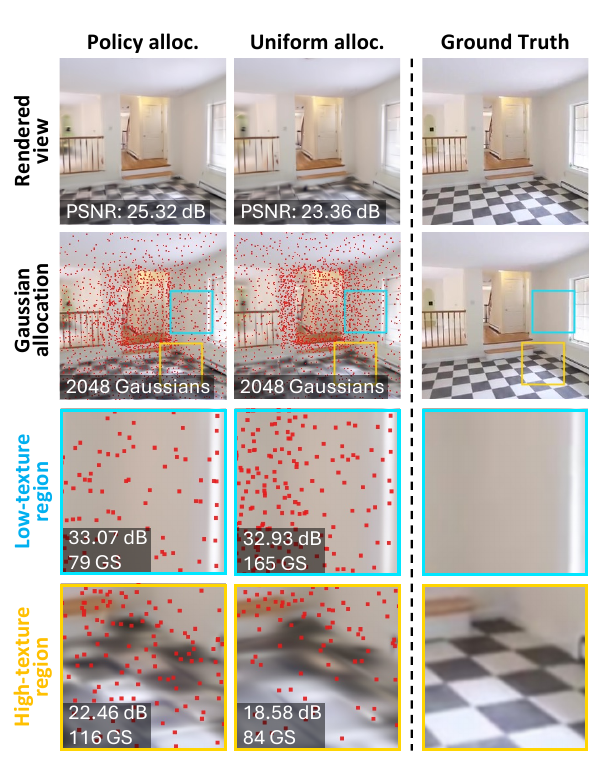}
        \vspace{-5mm}
        \caption{\textbf{Effect of the policy allocation.} At the same budget, the policy assigns fewer Gaussians than uniform allocation to the low-texture region (79 vs.\ 165) at similar PSNR, and more to the high-texture region (116 vs.\ 84), improving detail reconstruction.}
        \label{fig:allocation_effect}
    \end{minipage}
    \hspace{0.01\textwidth}
    \begin{minipage}[t]{0.46\columnwidth}
        \centering
        \includegraphics[trim={0 0 -10pt 0}, clip, width=\linewidth]{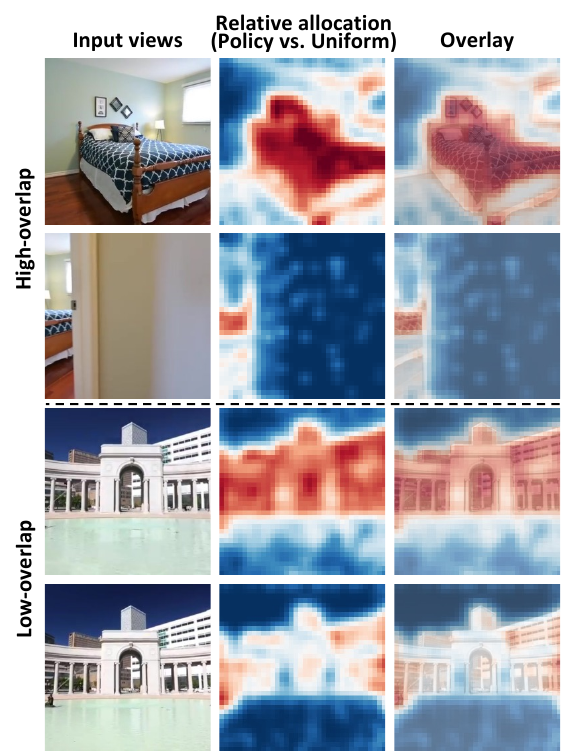}
        \vspace{-5mm}        
        \caption{\textbf{Learned allocation policy.} Two RE10K scenes, each with two input views and 1024 Gaussians per view on average. Maps show expected anchor allocation relative to uniform (red: more, blue: fewer), smoothed for display.
        }
        \label{fig:policy}
    \end{minipage}
    \vspace{-4mm}
\end{figure}

\begin{wraptable}[10]{r}{0.37\textwidth}
\centering
\footnotesize
\setlength{\tabcolsep}{0.15cm}
\vspace{-6mm}
  \caption{\textbf{Allocation strategy ablation.}
$^\dagger$Base model trained with random allocations, before fine-tuning.}
  \label{tab:ablation_alloc}
  \begin{tabular}{lccc}
    \toprule
    Variant & LPIPS $\downarrow$ & SSIM $\uparrow$ & PSNR $\uparrow$ \\
    \midrule
    Base$^\dagger$ & 0.277 & 0.708 & 24.18 \\
    \midrule
    Uniform        & 0.271 & 0.720 & 24.53 \\
    Edge-based     & 0.268 & 0.727 & 24.76 \\
    \textbf{Policy}  & \textbf{0.253} & \textbf{0.734} & \textbf{24.83} \\
    \bottomrule
  \end{tabular}
\end{wraptable}
\noindent\textbf{Allocation policy.}
\Cref{tab:ablation_alloc} compares our Gaussian allocation policy with learning-free alternatives, all fine-tuned from the same base-trained model: a uniform split and an edge-based heuristic that allocates Gaussians in proportion to the Sobel gradient magnitude of the image. The policy-based allocation achieves the best result on all three metrics.
\Cref{fig:allocation_effect} shows where this gain comes from. At the same total budget, the policy places fewer Gaussians than uniform allocation in a low-texture wall region with comparable rendering quality, and spends the saved budget on the textured floor, where it recovers the checkerboard pattern that uniform allocation blurs.
\Cref{fig:policy} visualizes the learned allocation over whole views. Across views, the policy moves budget away from the near-textureless view; within a view, it concentrates Gaussians on complex structures such as the bed and the building facade, while nearly emptying flat walls, floor, and water.

\noindent\textbf{Gaussian local refinement.} 
\Cref{tab:knn_cost} compares our hierarchical neighbor search with an exact search over all $B$ Gaussians at 65,536 Gaussians, using $K_{\rm{anc}}{=}16$ anchor neighbors and $K_{\rm{gs}}{=}16$ Gaussian neighbors. The hierarchical search is $27\times$ faster (8.1\,ms vs.\ 219.6\,ms) and reduces the end-to-end latency from 595.0\,ms to 388.0\,ms. With the Voronoi loss, the hierarchical search retrieves exactly the same $K_{\rm{gs}}$ nearest neighbors as the exact search for 99.0\% of the Gaussians, whereas without it this rate drops to 60.2\%, as Gaussians drift out of their anchor's cell.
\Cref{fig:topk_graph} varies the number of neighbors $K_{\rm{gs}}$ at inference while keeping $K_{\rm{anc}}$ fixed. For every $K_{\rm{gs}}$, peak VRAM decreases linearly with $B$, and a larger $K_{\rm{gs}}$ improves quality at the cost of memory. As the budget grows, the quality gap between different $K_{\rm{gs}}$ narrows, while the memory cost of a larger $K_{\rm{gs}}$ increases. Since $K_{\rm{gs}}$ can be changed without retraining, users can choose a smaller value at large budgets to save memory with little loss in quality, and a larger $K_{\rm{gs}}$ at small budgets where it matters most.

\begin{table}[H]
\begin{minipage}[t]{0.38\linewidth}
\vspace{0pt}
  \centering
  \small
  \setlength{\tabcolsep}{0.10cm}
  \caption{\textbf{Neighbor search cost.} Comparison of exact and hierarchical neighbor search at 65,536 Gaussians, reporting both search latency and end-to-end generation latency.}
  \label{tab:knn_cost}
  \begin{tabular}{lcc}
    \toprule
    \multirow{2}{*}{Neighbor search}
      & \multicolumn{2}{c}{Latency (ms) $\downarrow$} \\
    \cmidrule(lr){2-3}
      & Search & End-to-end \\
    \midrule
    Exact               & 220.8 & 539.2 \\
    Hierarchical (ours) & \textbf{8.1} & \textbf{328.3} \\
    \bottomrule
  \end{tabular}
\end{minipage}
\hfill
\begin{minipage}[t]{0.60\linewidth}
\vspace{-10pt}
  \centering
  \includegraphics[width=\linewidth]{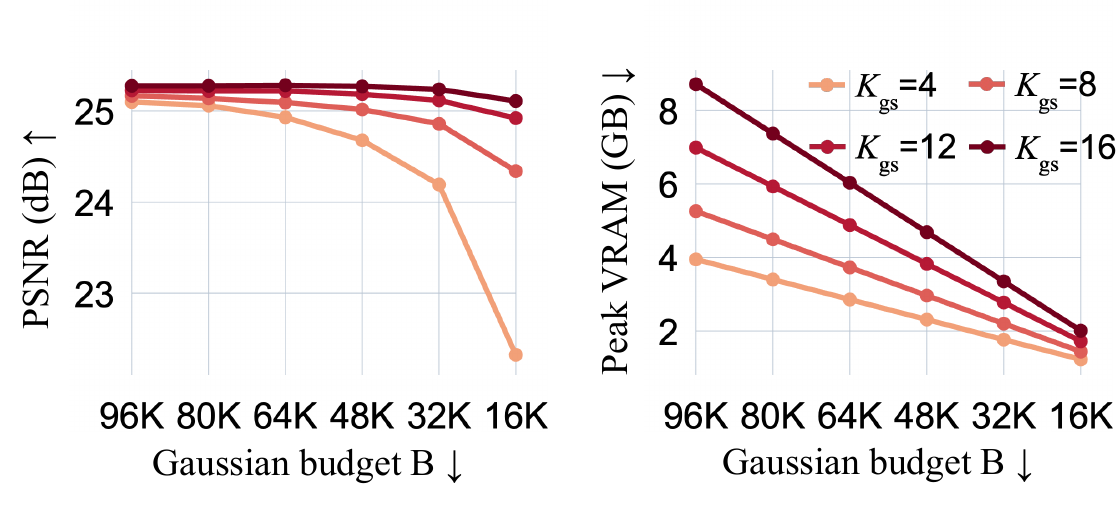}
  \vspace{-14pt}
  \captionof{figure}{
  \textbf{Gaussian-neighbor ablation.}
   $K_{\rm gs}$ is varied at inference with a single trained model; \(K_{\rm anc}{=}16\) is fixed.
  }
  \label{fig:topk_graph}
\end{minipage}
\vspace{-4pt}
\end{table}
\vspace{-7mm}
\section{Conclusion}
\vspace{-2mm}
We presented RLSplat, a budget-controllable feed-forward 3DGS framework whose inference cost scales linearly with the requested number of Gaussians $B$. A Gaussian allocation policy, trained with GSPO, distributes exactly $B$ Gaussians across anchors, and a resource-linear decoder refines them through local cross-attention with a hierarchical neighbor search. On ACID, \OURMODEL achieves the best rendering quality among budget-controllable methods at every tested budget. Its latency and peak memory decrease linearly with $B$, and it matches or exceeds the quality of the baselines at a fraction of their memory and latency.

\newpage

\bibliography{iclr2027_conference}
\bibliographystyle{iclr2027_conference}

\appendix
\newpage
\section{Limitations and Future Work} \label{supp:limitations} 
Image encoding remains a fixed cost independent of $B$, which bounds the achievable speed-up at small budgets. We also evaluate at a resolution of $256\times256$ and in a single cross-dataset setting. RLSplat controls both the size of the representation and the cost of producing it. It can therefore serve as a 3D scene representation for downstream tasks that have their own resource limits, such as 3D vision-language models and robot learning. Evaluating it on such tasks is a natural next step.
\section{Allocation Policy Details}
\label{supp:policy}

\paragraph{Policy head.}
The policy head predicts a logit $l_n$ for each anchor token, and the
allocation probabilities are obtained as
$\mathbf{p} = \operatorname{softmax}(\mathbf{l})$ with
$\mathbf{l} = \{l_n\}_{n=1}^N$. The head is
$\mathrm{LN}(d) \rightarrow \mathrm{Linear}(d, h) \rightarrow
\mathrm{SiLU} \rightarrow \mathrm{Linear}(h, 1)$, applied per anchor token.
The last linear layer is zero-initialized, so $\mathbf{p}$ starts uniform.

\paragraph{Budget conditioning.}
The head is modulated by FiLM parameters $(\boldsymbol{\gamma},\boldsymbol{\eta})$. They are predicted from a single scalar, the log-ratio of the requested budget to the number of anchors, $u=\log_2(B/N)$, by $\mathrm{Linear}(1,h)\rightarrow\mathrm{SiLU}\rightarrow\mathrm{Linear}(h,2d)$. The modulation is applied after the layer norm: $\text{hidden}\leftarrow\text{hidden}\odot\big(1+\tanh\boldsymbol{\gamma}(u)\big)+\boldsymbol{\eta}(u)$. The last layer of the FiLM generator is also zero-initialized. The ratio $B/N$ is floored at $1/64$ so that $u$ stays finite for very small budgets.

\paragraph{Policy initialization at the start of policy training.}
The policy head does not exist during base training. It is created with zero-initialized output layers when policy training begins, so it outputs exactly $0$ for every anchor and $\mathbf{p}$ is uniform at the first policy-training step. Base training's random allocation already covers this regime, so the transition introduces no distribution shift for the decoder.

\paragraph{Deterministic integer allocation.}
At inference, and for the reconstruction loss during policy training, the expected counts $B\mathbf{p}$ are converted into an integer allocation $\mathbf{k}^{\star}$ by systematic rounding with a fixed phase. Anchors are laid out on a line in a fixed (view, patch-raster) order, anchor $n$ owning the interval $[BP_{n-1},\,BP_n)$ with $P_n=\sum_{m\le n}p_m$ and $P_0=0$. We place $B$ points at $\varphi,\varphi+1,\dots,\varphi+B-1$ with $\varphi=\tfrac12$ and count those that fall inside each interval:
\begin{equation}
  k^{\star}_n \;=\; \big\lceil BP_n-\varphi \big\rceil-\big\lceil BP_{n-1}-\varphi \big\rceil .
  \label{eq:systematic}
\end{equation}
This rule has three properties we rely on: (i)~the counts sum to exactly $B$, since the telescoping sum equals $\lceil B-\tfrac12\rceil-\lceil-\tfrac12\rceil=B$; (ii)~every count is either $\lfloor Bp_n\rfloor$ or $\lceil Bp_n\rceil$, since an interval of length $Bp_n$ contains that many unit-spaced points; and (iii)~an anchor with $p_n=0$ receives no Gaussian. With a uniformly random phase, \cref{eq:systematic} is systematic resampling and satisfies $\mathbb{E}[k_n]=Bp_n$, the mean of the multinomial rollouts. Fixing $\varphi$ makes it deterministic. It needs a single cumulative sum, so it takes $\mathcal{O}(N)$ time, involves no sorting and has no ties to break. We compute it in double precision and renormalize $P_N$ to one.



\section{GSPO Training Details}
\label{supp:gspo}


\paragraph{GSPO hyperparameters.}
\label{supp:gspo_hparams}
\Cref{tab:ablation_gspo} ablates the two hyperparameters of the rollouts: the group size $C$ and the rollout temperature $\tau$. With $\tau\neq1$, the rollouts are drawn from the tempered proposal $\mathbf{q}$ rather than from $\mathbf{p}$, and the ratio is no longer one. The policy is robust to both hyperparameters: all settings are within 0.11\,dB PSNR of each other. $C=5$ performs best, and $\tau=1.0$, which samples from the policy itself, outperforms both tempered variants. We use $C=5$ and $\tau=1.0$ throughout.

\begin{table}[ht]
\centering
\footnotesize
\caption{\textbf{GSPO hyperparameters.} Group size $C$ and rollout
temperature $\tau$. The first row is our default.}
\label{tab:ablation_gspo}
\begin{tabular}{ccccc}
\toprule
$C$ & $\tau$ & LPIPS $\downarrow$ & SSIM $\uparrow$ & PSNR $\uparrow$ \\
\midrule
5 & 1.0 & \textbf{0.268} & \textbf{0.722} & \textbf{24.68} \\
\midrule
3 & 1.0 & 0.273 & 0.717 & 24.57 \\
7 & 1.0 & 0.269 & 0.720 & 24.63 \\
5 & 0.8 & 0.271 & 0.718 & 24.60 \\
5 & 1.2 & 0.270 & 0.719 & 24.61 \\
\bottomrule
\end{tabular}
\end{table}

\paragraph{Settled hyperparameters.}
Policy training uses $C=5$ rollouts and policy loss weight
$\lambda_{\mathrm{policy}} = 1.0$; the entropy and KL weights are both $0$, and no minimum per-anchor count is enforced.
Base training (\cref{sec:training}) trains the encoder and decoder without an allocation head at all, so there is no pretrained policy to serve as a KL reference at the start of policy training. The policy instead starts from the zero-initialized, uniform head described above.
\section{Decoder Details}
\label{supp:decoder}

\paragraph{Gaussian token construction.}
For anchor $n$ with $k_n$ allocated Gaussians, token $r \in \{0,\dots,k_n-1\}$
has relative position $\rho_{n,r} = (r+\tfrac12)/k_n \in (0,1)$, so the
tokens of any anchor tile the unit interval regardless of $k_n$. We use a
mip-NeRF-style~\citep{barron2021mip} integrated positional encoding
$\mathrm{PE}(\rho, k)$ that averages Fourier features over the token's
bin of width $1/k_n$, which suppresses the frequencies that would
otherwise alias between nearby values of $k_n$. The token is
\begin{equation}
  \mathbf{t}^{0}_{n,r}
  = \mathbf{a}_n
  + \big(1 + \tanh \bm{\alpha}_n\big) \odot
    \mathrm{MLP}\big(\mathrm{PE}(\rho_{n,r}, k_n)\big)
  + \bm{\beta}_n,
  \qquad
  (\bm{\alpha}_n, \bm{\beta}_n) = \mathrm{Linear}\big(\mathrm{LN}(\mathbf{a}_n)\big),
  \label{eq:gaussian_token}
\end{equation}
with the FiLM layer zero-initialized. All $B$ tokens across a batch are
packed into one sequence, so no computation is spent on padding.

\paragraph{Neighbor search.}
Anchor centroids $\bar{\bm{\mu}}_n$ are the mean initial centers of
each anchor's tokens. $\mathcal{N}^{\mathrm{anc}}(n)$, the $K_{\rm{anc}} {=} 16$ nearest anchors
to $n$ (including $n$ itself), is found once by a $K$-nearest-neighbor
search over the detached centroids and shared by all $L=2$ decoder
layers. A Gaussian's neighbor set $\mathcal{N}^{\mathrm{gs}}(b)$ is the $K_{\rm{gs}}$ nearest Gaussian tokens, by detached Euclidean distance, among the tokens belonging to $\mathcal{N}^{\mathrm{anc}}(n_b)$.

\paragraph{Residual head.}
The initial-Gaussian head and the residual head share the same
architecture, $\mathrm{LN} \rightarrow \mathrm{Linear}(d, 256)
\rightarrow \mathrm{SiLU} \rightarrow \mathrm{Linear}(256, 11 + 3d_{\mathrm{sh}})$,
predicting a center offset (3), an opacity logit (1), scales (3), a
quaternion (4), and spherical-harmonic color coefficients. The residual
head is zero-initialized, and its center output is scaled by
$\alpha_{\mu}$.

\section{Training Details}
\label{supp:training}

Context views are loaded at $256\times256$ and bilinearly upsampled to $504\times504$ immediately before the backbone, while target views, losses, and evaluation remain at $256\times256$. The DVLT~\citep{burzio2026dvlt} backbone runs 12 inference steps. Each training step draws the number of context views $V$ from the divisors of 24, $V \in \{2, 3, 4, 6, 8, 12, 24\}$, and sets the batch size to $24/V$, so that every step processes 24 context views in total regardless of $V$.

\paragraph{Base training.}
The encoder and decoder are trained for 40K steps with allocations drawn uniformly at random, which exposes the decoder to a wide range of per-anchor counts before any policy is learned. The per-sample budget is sampled from a Beta distribution over $[2048, 16384V]$. We use AdamW with a learning rate of $10^{-4}$ ($0.1\times$ for the backbone), 1,000 warm-up steps, and cosine decay on 4 GPUs. The loss weights are $1.0$ for MSE, $0.05$ for VGG-LPIPS, $0.1$ for the reprojection term and $0.05$ for the cheirality term of $\mathcal{L}_{\mathrm{proj}}$, and $1.0$ for $\mathcal{L}_{\mathrm{vor}}$. The offset loss $\mathcal{L}_{\mathrm{offset}}$ is weighted $\lambda_{\mathrm{offset}} = 0.01$.

\paragraph{Policy training.}
Starting from the pretrained model, all parameters are fine-tuned for 5K
steps with $C=5$ rollouts per step and a rollout temperature of
$\tau=1.0$; the render used for the reconstruction loss is produced from the
deterministic allocation of \cref{sec:policy}. The learning rate is reduced to
$10^{-5}$, with a $30\times$ multiplier on the allocation head, and
$\mathcal{L}_{\mathrm{offset}}$ is disabled. The policy loss
$\mathcal{L}_{\mathrm{policy}}$ is weighted $1.0$. Each sample's budget is drawn independently from $\mathrm{Beta}(1,4)$
over the same range, which biases training toward low budgets where allocation
matters most. The total budget of a step is capped, so that peak memory is
bounded independently of the per-sample maximum.

\section{Ablation Protocol}
\label{supp:ablation}

Each ablation changes a single component relative to its reference, except for the variant that removes both locality regularizers. \emph{Pretraining ablations} (\cref{tab:ablation_arch}) are trained from scratch for 5K steps, one eighth of the full schedule, with all step-dependent schedules (cosine horizon, optimizer warm-up, view-sampler warm-up, and validation and checkpoint intervals) scaled by the same factor. We compare the full recipe with variants that remove the local refinement module, the projection loss (both its reprojection and cheirality terms), the offset loss, or both the offset and Voronoi losses. The neighbor match rate without the Voronoi loss is measured with the same schedule. \emph{Allocation ablations} (\cref{tab:ablation_alloc,tab:ablation_gspo}) fine-tune for 1K steps from a single shared pretrained checkpoint and vary the group size $C \in \{3, 5, 7\}$, the temperature $\tau \in \{0.8, 1.0, 1.2\}$, and the allocation rule, replacing the learned policy with a uniform split or a learning-free Sobel-energy heuristic. Since a shorter run anneals on its own cosine schedule, each variant is trained for its own horizon rather than truncated from a longer run. All ablations are evaluated on ACID with 8 context views (249 scenes). Allocation ablations are evaluated at $B/V=1024$.

\section{Allocation Policy Visualizations}
\label{supp:policy_vis}

\begin{figure}[p]
  \centering
  \includegraphics[width=0.95\textwidth]{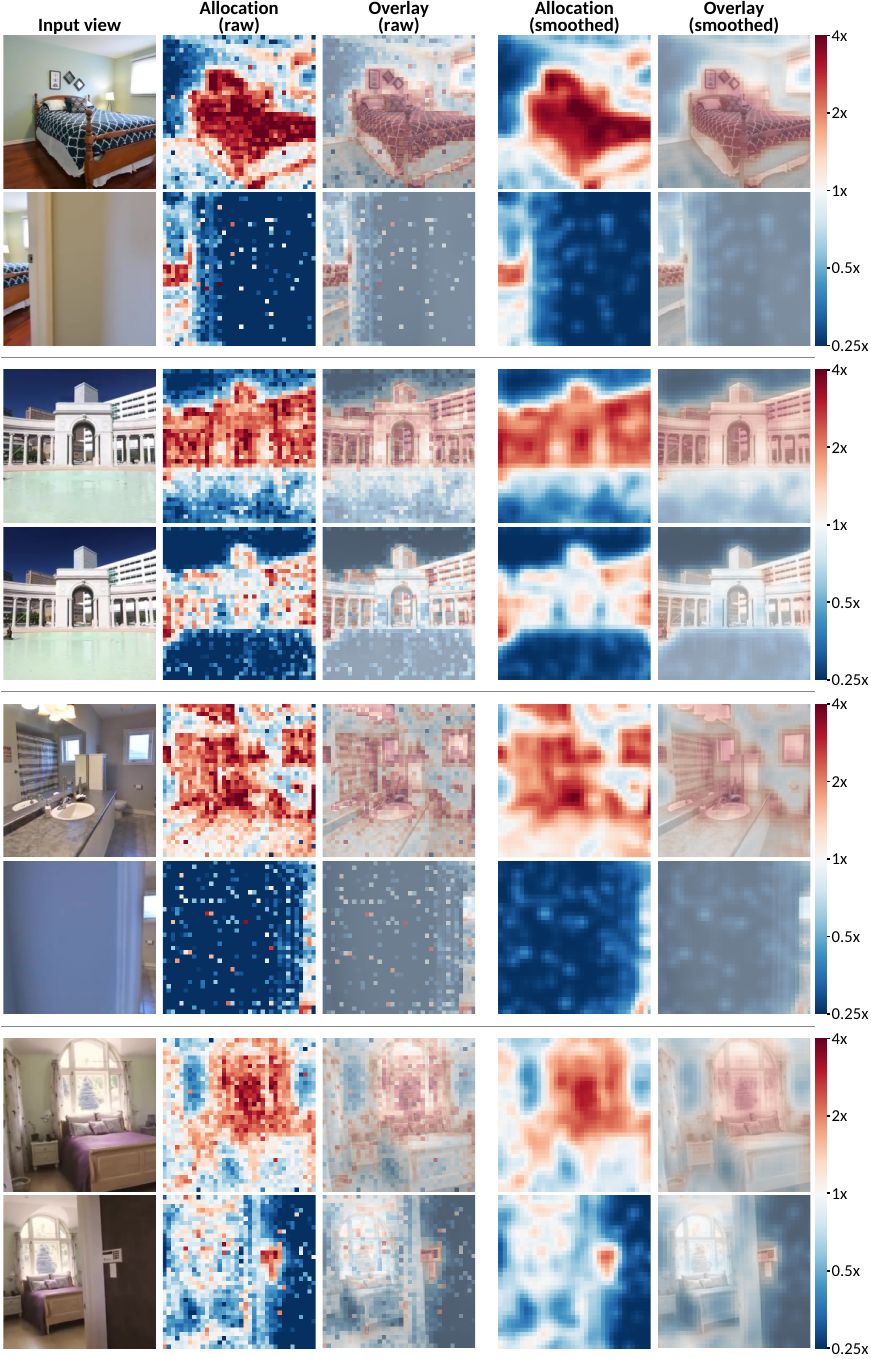}
  \vspace{-5mm}
  \caption{\textbf{More allocation policies (1/3).}
  Raw (left pair) and smoothed (right pair) allocation maps on a log scale
  ($0.25\times$ to $4\times$; red: more, blue: fewer). In these indoor
  scenes the policy shifts budget toward the view with more structure and
  away from near-blank walls and doorways.}
  \label{fig:supp_policy_1}
\end{figure}

\begin{figure}[p]
  \centering
  \includegraphics[width=0.95\textwidth]{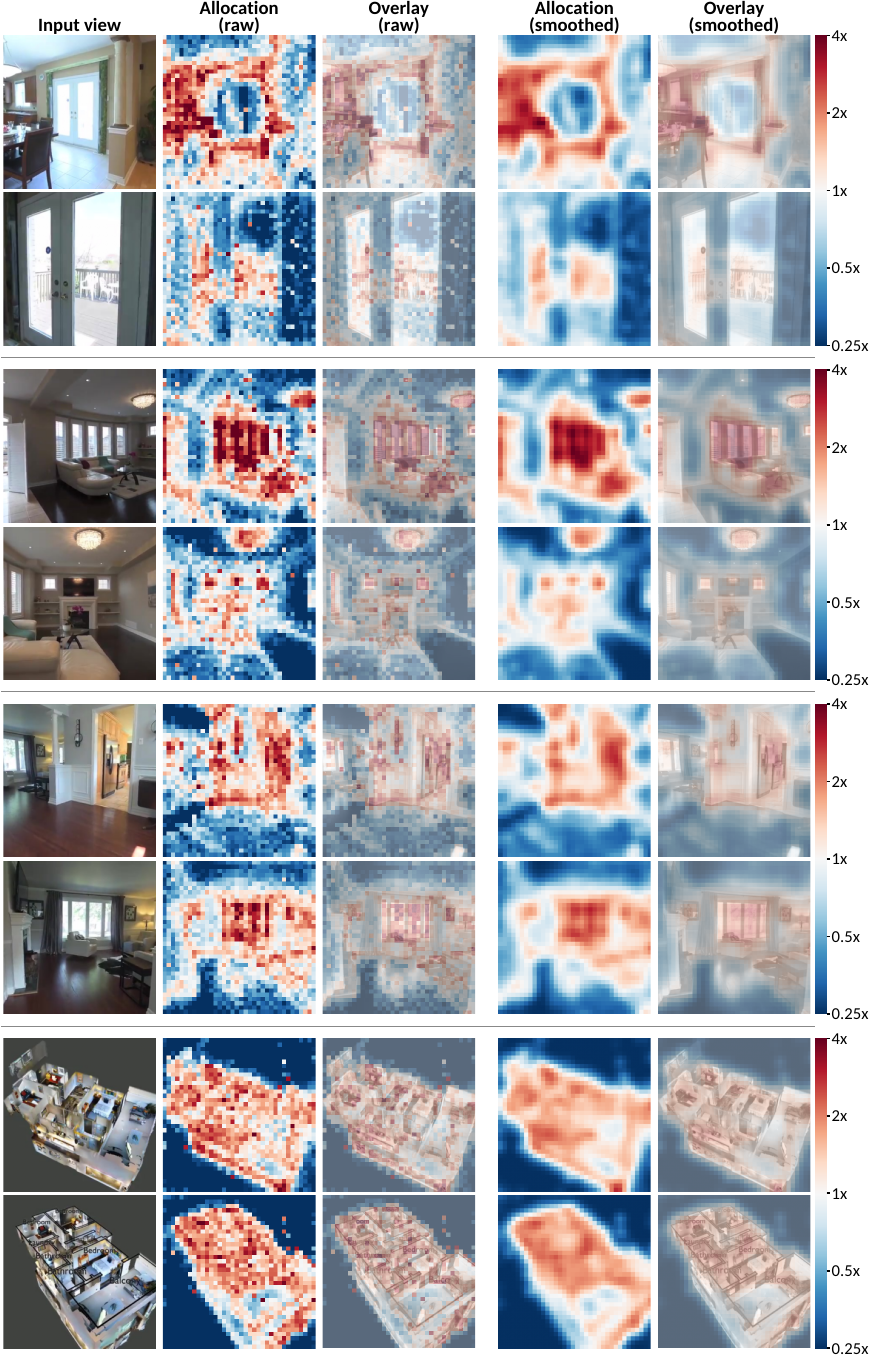}
  \vspace{-5mm}
  \caption{\textbf{More allocation policies (2/3).}
  Same layout as \cref{fig:supp_policy_1}. Budget moves away from dark or
  under-exposed regions, such as the arch frame and the darkened hillside,
  and concentrates on the lit part of the scene or on the object.}
  \label{fig:supp_policy_2}
\end{figure}

\begin{figure}[p]
  \centering
  \includegraphics[width=\textwidth]{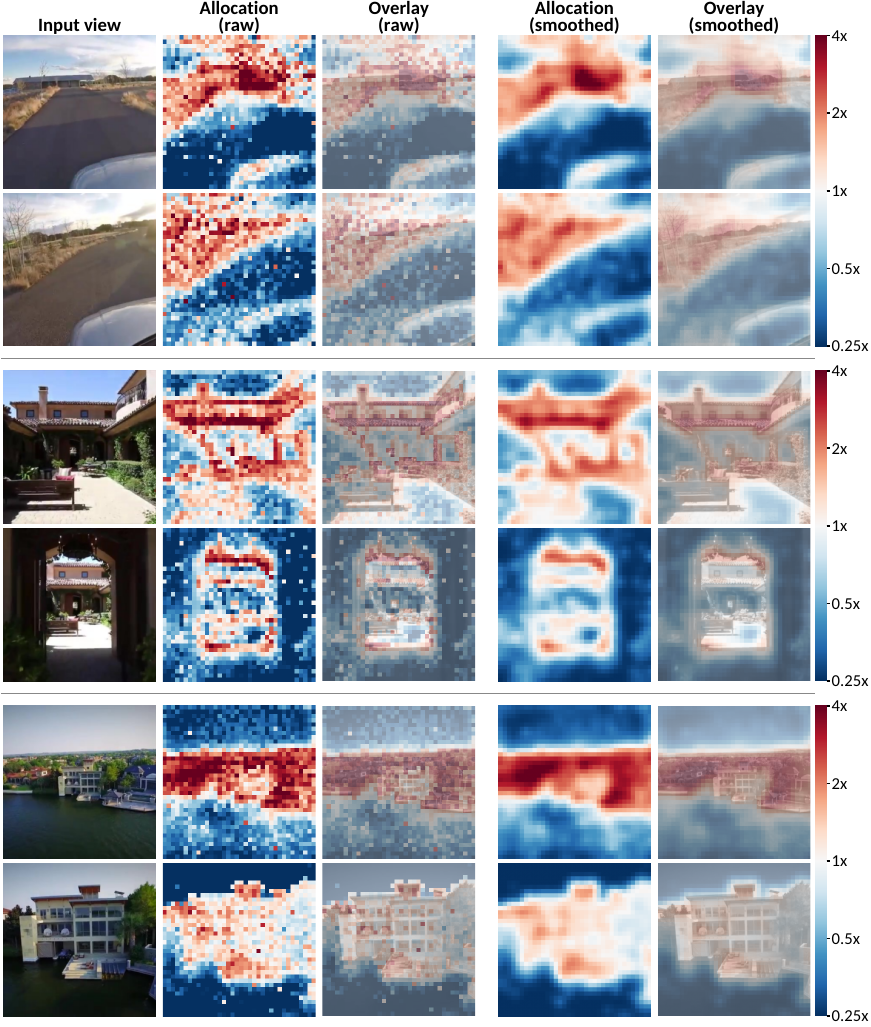}
  \caption{\textbf{More allocation policies (3/3).}
  Same layout as \cref{fig:supp_policy_1}. When the two views are nearly
  identical, the budget stays close to an even split between them, and the
  policy acts within each view instead: it empties water and sky and spends
  the budget on buildings, vegetation and shorelines.}
  \label{fig:supp_policy_3}
\end{figure}

\cref{fig:supp_policy_1,fig:supp_policy_2,fig:supp_policy_3} extend \cref{fig:policy} to twelve RE10K two-view test scenes, all at an average of 1024 Gaussians per view. For each anchor we show its expected allocation relative to a uniform split, $p_n N$, where $N$ is the number of anchors over both views; $1\times$ corresponds to about $0.8$ Gaussians per anchor at this budget. This is the policy's output before the budget is rounded to integer counts. The corner label is the fraction of the total budget that goes to that view. We show the raw per-anchor maps next to the smoothed maps used in the main paper (Gaussian filter, $\sigma{=}1$ anchor, applied to $\log_2$ of the map), so the smoothing can be checked for artifacts: it only removes anchor-level noise and leaves the spatial pattern unchanged.

\section{Additional Qualitative Comparison}
\label{supp:qual_vis}
\begin{figure}[p]
  \centering
  \includegraphics[width=\textwidth]{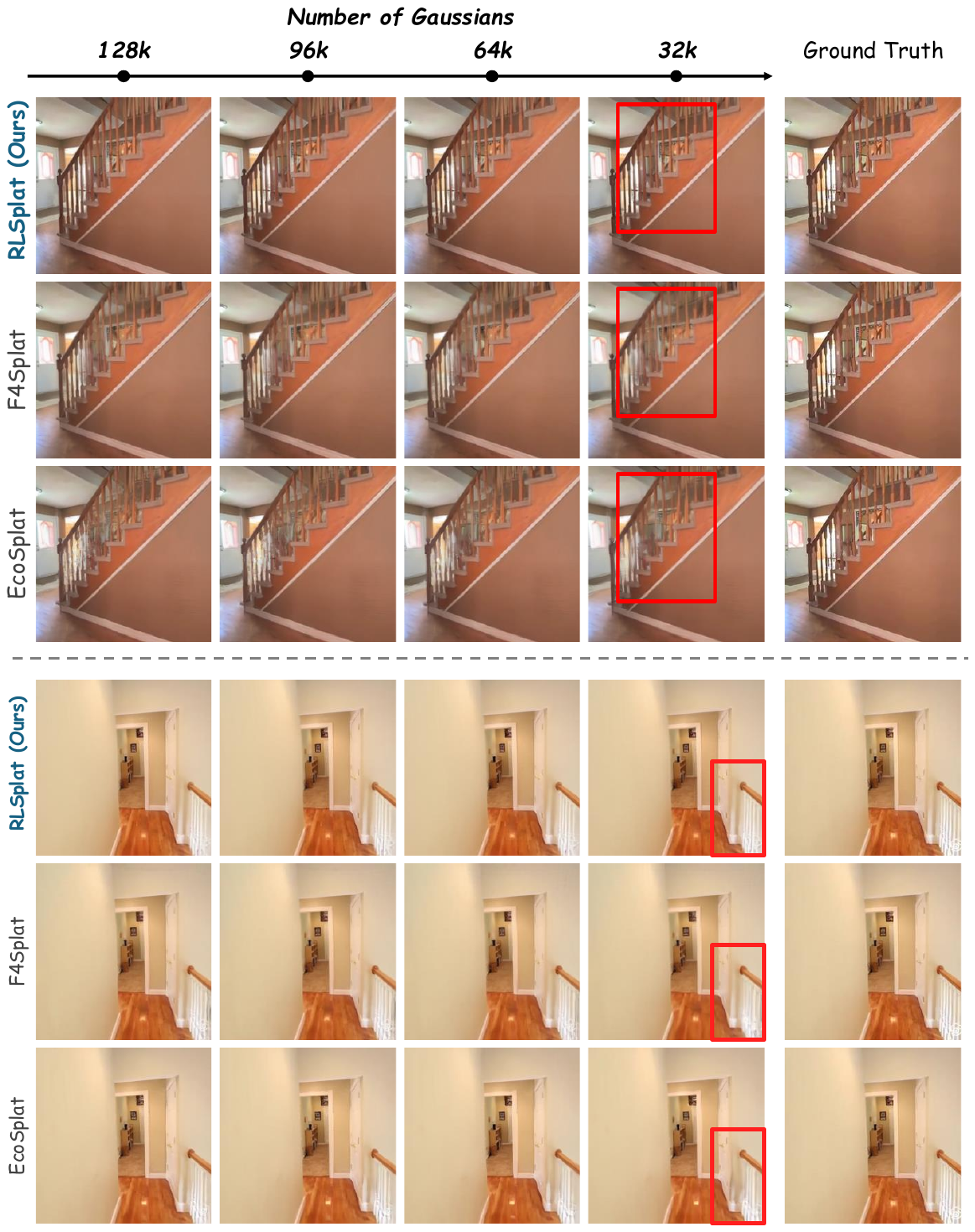}
\vspace{-20mm}
  \caption{\textbf{Qualitative comparison across Gaussian budgets (1/3).}}
  \label{fig:supp_qual_comp_1}
\end{figure}

\begin{figure}[p]
  \centering
  \includegraphics[width=\textwidth]{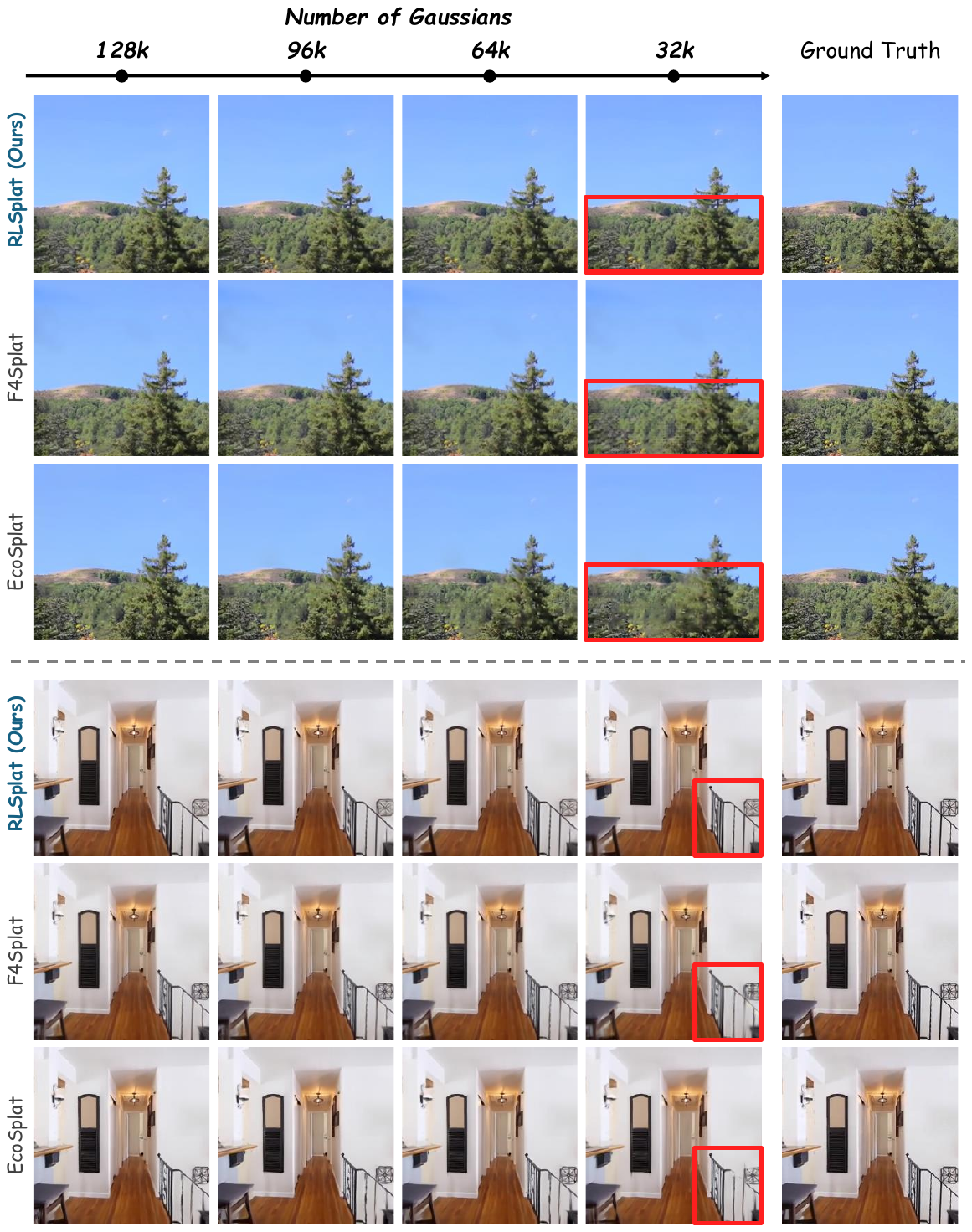}
\vspace{-20mm}
  \caption{\textbf{Qualitative comparison across Gaussian budgets (2/3).}}
  \label{fig:supp_qual_comp_2}
\end{figure}

\begin{figure}[p]
  \centering
  \includegraphics[width=\textwidth]{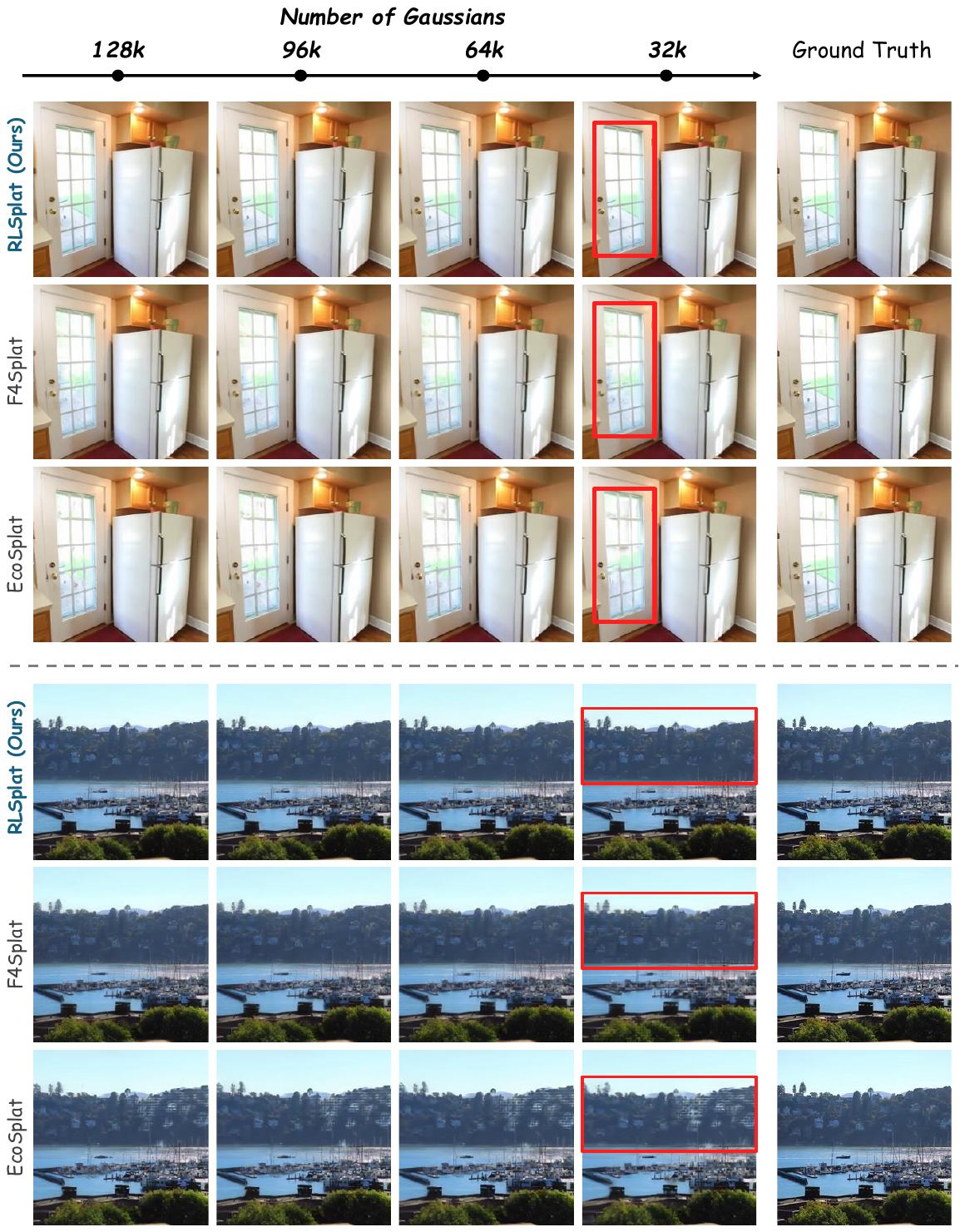}
\vspace{-20mm}  
  \caption{\textbf{Qualitative comparison across Gaussian budgets (3/3).}}
  \label{fig:supp_qual_comp_3}
\end{figure}

\cref{fig:supp_qual_comp_1,fig:supp_qual_comp_2,fig:supp_qual_comp_3} compare RLSplat with F$^4$Splat and EcoSplat as the Gaussian budget is reduced from 128K to 32K.
As the budget decreases, the baselines progressively blur or lose thin, high-frequency structures such as stair balusters, railings, window frames, and foliage, while RLSplat retains these details at every budget.
This is consistent with \cref{tab:acid_budget_controllable}, where RLSplat exhibits substantially less quality degradation than the baselines as the budget decreases.
Unlike the baselines, which reduce a dense set of candidates, RLSplat allocates the budget where the scene needs it and decodes only those Gaussians, preserving fine details even under the tightest budgets.

\end{document}